\pdfoutput=1 
\documentclass[11pt,a4paper]{article}
\usepackage[hyperref]{ranlp2023}
\usepackage{booktabs}
\usepackage{times}
\usepackage{tabularx}
\usepackage{latexsym}
\usepackage{makecell}
\usepackage{amssymb}
\usepackage[nolist]{acronym}

\begin{acronym}
    \acro{nlp}[NLP]{natural language processing}
    \acro{llm}[LLM]{large language model}
    \acroplural{llm}[LLMs]{large language models}
\end{acronym}

\usepackage{microtype}

\newcommand{\y}[0]{\checkmark}
\newcommand{\n}[0]{\texttimes}

\aclfinalcopy 

\title{Detecting ChatGPT: A Survey of the State of Detecting ChatGPT-Generated Text}

\author{Mahdi Dhaini \hspace{4em} Wessel Poelman \hspace{4em} Ege Erdogan \\
        Technical University of Munich \\ Department of Computer Science \\ Germany \\
        \texttt{\{mahdi.dhaini,wessel.poelman,ege.erdogan\}@tum.de}}

\date{}

\begin{document}
\maketitle
\begin{abstract}
While recent advancements in the capabilities and widespread accessibility of generative language models, such as ChatGPT \cite{openaiChatGPTOptimizingLanguage2022}, have brought about various benefits by generating fluent human-like text, the task of distinguishing between human- and \ac{llm} generated text has emerged as a crucial problem.
These models can potentially deceive by generating artificial text that appears to be human-generated. This issue is particularly significant in domains such as law, education, and science, where ensuring the integrity of text is of the utmost importance.
This survey provides an overview of the current approaches employed to differentiate between texts generated by humans and ChatGPT.
We present an account of the different datasets constructed for detecting ChatGPT-generated text, the various methods utilized, what qualitative analyses into the characteristics of human versus ChatGPT-generated text have been performed, and finally, summarize our findings into general insights.
\end{abstract}

\section{Introduction}
\acp{llm} have been showing remarkable abilities in generating fluent, grammatical, and convincing text.
The introduction of ChatGPT \citep{openaiChatGPTOptimizingLanguage2022} has been widely regarded as a significant and controversial milestone for \acp{llm}. 
Models such as GPT-3 \citep{brownLanguageModelsAre2020} and PaLM \citep{chowdheryPaLMScalingLanguage2022} already demonstrated the power of \acp{llm} in many \ac{nlp} tasks.
ChatGPT is the first model that has seen widespread adoption outside NLP research.

The increased performance of \acp{llm} raises important questions regarding their potential societal impact. 
The risks of \acp{llm} are numerous, from confidently presenting false information to generating fake news on a large scale \cite{shengSocietalBiasesLanguage2021, weidingerTaxonomyRisksPosed2022}. 
ChatGPT is no exception in this regard \citep{zhuoRedTeamingChatGPT2023}.

Instances of the misuse of ChatGPT have already been documented in various domains, including education \cite{cottonChattingCheatingEnsuring2023}, scientific writing \cite{gaoComparingScientificAbstracts2022}, and the medical field \cite{andersonAIDidNot2023}. 
Given this context, the detection of machine-generated text is gaining considerable attention.
This detection is part of a larger push towards responsible and appropriate usage of generative language models \citep{kumarLanguageGenerationModels2023}.

In addition to academic interest, a growing number of commercial parties are trying to solve this task.
Recent work from \citet{pegoraroChatGPTNotChatGPT2023} gives an overview of commercial and freely available online tools.
They come close to the current work. However, we limit our scope to academic work and provide additional background information on methods, datasets, and qualitative insights.

Many approaches, datasets and shared tasks\footnote{For instance, \href{https://sites.google.com/view/autextification/home}{AuTexTification} or \href{https://sites.google.com/view/shared-task-clin33/home}{CLIN33 Shared Task}.} have been put forth recently to tackle the \mbox{\emph{general}} (i.e., not specific for ChatGPT) task of detecting machine-generated text \citep{jawaharAutomaticDetectionMachine2020}.
Given the enormous use and cultural impact of ChatGPT, we limit our review to datasets and methods developed directly for ChatGPT.
We discuss these methods in the context of the controversial position ChatGPT is in, namely that it is a closed-source system with very little information available regarding its training setup or model architecture at the time of writing.
We outline what general methods exists for this task and review recent work that directly focuses on datasets and methods for ChatGPT.

Given the peculiar place ChatGPT is in, we also consolidate qualitative insights and findings from the works we discuss that might help humans to detect ChatGPT-generated text.
These include linguistic features or writing styles to look out for.
Lastly, we present outstanding challenges for this detection task and possible future directions.

\section{Related Work on Detecting Machine-Generated Text}

\acp{llm} have become a driving force in many language processing-related benchmarks and tasks \citep{radfordLanguageModelsAre2018,brownLanguageModelsAre2020,chowdheryPaLMScalingLanguage2022}. \acp{llm} can solve complex NLP tasks and generate convincing and trustworthy-looking text.
However, they are also prone to generating false and misleading information, generally referred to as \emph{hallucinating} \citep{linTruthfulQAMeasuringHow2022}.
Additionally, misuse of these models can pose significant risks in academia, journalism, and many other areas.
Currently, human judges are decent at spotting machine-generated text from older \acp{llm} such as GPT-2 \citep{ippolitoAutomaticDetectionGenerated2020, duganRoFTToolEvaluating2020,duganRealFakeText2023}. Still, the increasing abilities of \acp{llm} give rise to the need for more sophisticated detection tools and models.

A recent survey by \citet{crothersMachineGeneratedText2023} provides a thorough overview of risks, approaches, and detection methods.
They discuss interesting aspects such as the effect of domains on the detection task, adversarial attacks, and societal impacts of generated texts.
Work done by \citet{jawaharAutomaticDetectionMachine2020} inspects the field of machine-generated text detection. 
It outlines three main detection methods: a classifier trained from scratch, zero-shot detection by a language model, and a fine-tuned language model as a classifier.
Recently, detection methods from computer vision have also been tried on language models, such as watermarking \citep{kirchenbauerWatermarkLargeLanguage2023,kirchenbauerReliabilityWatermarksLarge2023} or trying to find model-identifying artifacts in generated content \citep{tayReverseEngineeringConfigurations2020}.
To use and evaluate these methods, fine-grained access to the source model is required in training and inference time.
Both these preconditions are not the case with ChatGPT at the time of writing.

When discussing detection methods, an essential factor to consider is access to the log probability output of a model.
This is the probability distribution over the vocabulary of a model for the next token to be generated.
Numerous successful detection methods evaluate the average log probability per token combined with a threshold in a zero-shot setting \citep{gehrmannGLTRStatisticalDetection2019,ippolitoAutomaticDetectionGenerated2020,mitchellDetectGPTZeroShotMachineGenerated2023}.
This method is model agnostic and generally performs quite well.
At the time of writing, users of ChatGPT do not have access to these probabilities.
Without this access or knowledge about model internals, detection methods are limited to using just the generated text in a binary classification setting, with the options being \emph{human} or \emph{machine}.
These methods use simple classifiers trained on n-grams \cite{solaimanReleaseStrategiesSocial2019,ippolitoAutomaticDetectionGenerated2020} or fine-tuned pre-trained language models \cite{uchenduAuthorshipAttributionNeural2020, ippolitoAutomaticDetectionGenerated2020,zellersDefendingNeuralFake2020}.

Another group of detection tools we want to mention are the \emph{human-machine collaboration systems}, as \citet{jawaharAutomaticDetectionMachine2020} labels them.
These tools do not necessarily classify a passage directly but assist a human in making that decision.
The previously mentioned work by \citet{gehrmannGLTRStatisticalDetection2019} visualizes tokens in different colors, depending on where a given token ends up in the top-$k$ most probable tokens from the model.
This can also assist a human judge in spotting which part of a larger text might be machine-generated, such as possibly rephrased or copied sections for example.
As mentioned, this method requires access to output probabilities, so it is not usable for ChatGPT.
Another tool to help humans in the detection task is to outline the linguistic properties and characteristics of machine-generated text.
This was one of the main goals of the \emph{Real or Fake Text?} (RoFT) game created by \citet{duganRoFTToolEvaluating2020,duganRealFakeText2023}.
This game asked players to decide if a machine partially wrote a piece of text, and if yes, where the transition point from human to machine is in the text.
This resulted in a considerable dataset of annotations and indicators humans look for in detecting machine-generated text.

Another area of research that might help humans to make this decision is explainable AI.
As we will see, some papers we discuss use explainability methods, such as SHAP \citep{lundbergUnifiedApproachInterpreting2017}, in their approaches.
These methods help to better understand how detectors make their predictions.
Such methods can help provide insights on the input features that most contribute to a prediction, which, in turn, can facilitate analyses of the differences between human and ChatGPT writing styles.

As far as we know, the previously mentioned work by \citet{crothersMachineGeneratedText2023} and \citet{jawaharAutomaticDetectionMachine2020} come closest to ours.
They discuss detection methods and datasets but not ChatGPT.
The work from \citet{pegoraroChatGPTNotChatGPT2023} does mention ChatGPT, among other models, but  focuses mainly on online detection tools.

Our contributions are the following:
\begin{itemize}
    \item We provide an overview of \textit{general} approaches to machine-generated text detection.
    \item We outline research that specifically addresses the detection of ChatGPT-generated text and how this relates to the general approaches.
    \item We show the datasets that are created and used for this detection task.
    \item We summarize the qualitative analyses that these recent works provide and try to give general insights.
\end{itemize}

\section{Review of Approaches for Detecting ChatGPT-Generated Text}

\begin{table*}[t!]
    \centering
    \begin{tabular}{lllll}
        \toprule
        \thead{Dataset (name)} & \thead{Domain} & \thead{Public} & \thead{OOD} & \thead{Size and Setup}   \\
        \midrule
        \citealt{guoHowCloseChatGPT2023} (HC3-English)                   & Multi-domain       & 
            \href{https://github.com/Hello-SimpleAI/chatgpt-comparison-detection}{\y}    & \n & 
            \makecell[l]{\emph{Q\&A} \\ Questions: 24,322 \\ Human-A: 58,546 \\ ChatGPT-A: 26,903} \\ \midrule
        \citealt{guoHowCloseChatGPT2023} (HC3-Chinese)                   & Multi-domain       & 
            \href{https://github.com/Hello-SimpleAI/chatgpt-comparison-detection}{\y}    & \n & 
            \makecell[l]{\emph{Q\&A} \\ Questions: 12,853 \\ Human-A: 22,259\\ ChatGPT-A: 17,522} \\ \midrule
        \citealt{yuCHEATLargescaleDataset2023} (CHEAT)                  & Scientific    & 
            \n         & \y &
            \makecell[l]{\textit{Abstracts} \\ Human: 15,395 \\ ChatGPT: 35,304} \\ \midrule
         \citealt{heMGTBenchBenchmarkingMachineGenerated2023} (MGTBench)  & General            & 
            \href{https://github.com/xinleihe/MGTBench}{\y}         & \n &
            \makecell[l]{\textit{Q\&A pairs} \\ Human: 2,817 \\ ChatGPT: 2,817} \\ \midrule
        \citealt{liuArguGPTEvaluatingUnderstanding2023} (ArguGPT)        & Education          & 
            \href{https://github.com/huhailinguist/ArguGPT}{\y}        & \n &
            \makecell[l]{\textit{Essays} \\ Human: 4,115 \\ ChatGPT: 4,038} \\ \midrule
        \citealt{vasilatosHowkGPTInvestigatingDetection2023}           & Education          & 
            \href{https://github.com/comnetsAD/ChatGPT}{Human*} & \n &
            \makecell[l]{\textit{Q\&A}\\ Questions: 320 \\ Human-A: 960 \\ ChatGPT-A: 960} \\ \midrule
        \citealt{mitrovicChatGPTHumanDetect2023}                       & \makecell[l]{General} & 
            \href{https://www.kaggle.com/competitions/restaurant-reviews/overview}{Human*} & \y &
            \makecell[l]{\textit{Reviews} \\ Human: 1,000 \\  ChatGPT-query: 395 \\ChatGPT-rephrase: 1,000}  \\ \midrule 
        \citealt{wengUnderstandingExplanationMixedInitiative2023}      & \makecell[l]{Scientific}         & 
            \href{https://figshare.com/articles/dataset/VitaLITy_A_Dataset_of_Academic_Articles/14329151}{Human} & \n &
            \makecell[l]{\textit{Title-Abstract pairs} \\ Human: 59,232 \\ ChatGPT: 59,232  } \\ \midrule
        \citealt{antounRobustDetectionLanguage2023}  & General & 
            \href{https://gitlab.inria.fr/wantoun/robust-chatgpt-detection}{\y} & \y & 
            \makecell[l]{\textit{Q\&A} \\ HC3-English \\ OOD-ChatGPT: 5,969}\\  \midrule
        \citealt{liaoDifferentiateChatGPTgeneratedHumanwritten2023} & Medical & 
            \href{https://www.kaggle.com/datasets/chaitanyakck/medical-text}{Human} & \n & 
            \makecell[l]{\textit{Abstracts and records} \\ Human: 2,200 \\ ChatGPT: 2,200} \\
        \bottomrule
    \end{tabular}
    \caption{Datasets used in ChatGPT-generated text detection, with public availability information (if a dataset is available, it can be accessed by clicking on its \textit{Public} column entry). The \textit{Human} entry in the \textit{Public} column signals that only human-written text datasets are made public. The \textit{OOD} (out-of-domain) column signals if a dataset contains examples generated in a different way than the main part (e.g., rephrasing of human-written text). *Authors state it will be made available at a future date.}
    \label{tab:datasets}
\end{table*}

\subsection{Datasets}

Table \ref{tab:datasets} shows datasets that can be used to perform analyses or train models to distinguish between human and ChatGPT written text. We describe how they were collected and provide further information on their domains and public availability. 

\subsubsection{\citealt{guoHowCloseChatGPT2023} (HC3)}

Available in both Chinese and English, the Human ChatGPT Comparison Corpus (HC3) contains question-answer pairs collected from different datasets such as OpenQA \cite{yangWikiQAChallengeDataset2015} and Reddit ELI5 \cite{fanELI5LongForm2019}. 
These questions are then given to ChatGPT with context-sensitive prompts (e.g., asking ChatGPT to answer \emph{like I am five} for the Reddit ELI5 dataset) so that each question has one human-generated and one ChatGPT-generated answer. 

\subsubsection{\citealt{yuCHEATLargescaleDataset2023} (CHEAT)}

The ChatGPT-written Abstract (CHEAT) dataset contains human- and ChatGPT-generated title-abstract pairs for computer science papers, with the titles and human-written abstracts fetched from IEEE Xplore. Artificial abstracts are generated in three ways:
\begin{itemize}
    \item \textit{Generate}: ChatGPT is directly prompted to write an abstract given the title and keywords. 
    \item \textit{Polish}: ChatGPT is given human-written abstracts and is told to ``polish'' them.
    \item \textit{Mix}: Text from human-written and polished abstracts are mixed at the sentence level. 
\end{itemize}
The CHEAT dataset also covers adversarial scenarios as the \textit{Polish} and \textit{Mix} methods correspond to methods a malicious user might try to evade detection.

\subsubsection{\citealt{heMGTBenchBenchmarkingMachineGenerated2023} (MGTBench)}

The Machine Generated Text Benchmark (MGTBench) uses three question-answering datasets: TruthfulQA \cite{linTruthfulQAMeasuringHow2022}, SQuaD1 \cite{rajpurkarSQuAD1000002016}, and NarrativeQA \cite{kociskyNarrativeQAReadingComprehension2018}.
Questions are randomly sampled from each dataset, and ChatGPT is prompted to answer them with the appropriate context (e.g., with a relevant passage and instructions for NarrativeQA). 

Although our primary focus is ChatGPT, MGTBench contains text generated by different language models and thus can be used to benchmark detection methods across models. 

\subsubsection{\citealt{liuArguGPTEvaluatingUnderstanding2023} (ArguGPT)}

The ArguGPT dataset contains prompts and responses from various English learning corpora, such as WECCL \cite{zhi-jiaReviewSpokenWritten2008}, TOEFL11 \cite{blanchardTOEFL11CORPUSNONNATIVE2013}, and hand-picked from graduate record examinations (GRE) preparation material.
The texts are from essay writing assignments about a given topic or standpoint.
GPT models are prompted to write responses, but their output is processed for grammatical errors and to remove obvious signs of ChatGPT-generated text (e.g., ``As a large language model\ldots'').

\subsubsection{\citealt{vasilatosHowkGPTInvestigatingDetection2023}}

The dataset used in \citet{vasilatosHowkGPTInvestigatingDetection2023} for detection builds on \citet{ibrahimPerceptionPerformanceDetectability2023}, a dataset of questions with metadata and student answers from various university courses. ChatGPT is directly prompted with the questions three times to obtain three human and ChatGPT answers for each question.

\subsubsection{\citealt{mitrovicChatGPTHumanDetect2023}}

Attempting to build a classifier to detect ChatGPT-generated restaurant reviews, \citet{mitrovicChatGPTHumanDetect2023} build on the Kaggle restaurant reviews dataset\footnote{\url{https://www.kaggle.com/competitions/restaurant-reviews/overview}} and prompt ChatGPT to generate reviews of various kinds (e.g., ``write a review for a bad restaurant''). Additionally, ChatGPT is prompted to rephrase the human-written reviews to create an adversarial set. 

\subsubsection{\citealt{wengUnderstandingExplanationMixedInitiative2023}}

\citet{wengUnderstandingExplanationMixedInitiative2023} expand on \citet{narechaniaVitaLITyPromotingSerendipitous2022}'s dataset of title-abstract pairs fetched from top data visualization venues by prompting ChatGPT to write abstracts given the titles. Compared to another dataset of title-abstract pairs, CHEAT \cite{yuCHEATLargescaleDataset2023}, \citet{wengUnderstandingExplanationMixedInitiative2023}'s dataset contains more examples but lacks the adversarial samples included in CHEAT.

\subsubsection{\citealt{antounRobustDetectionLanguage2023}}

\citet{antounRobustDetectionLanguage2023} extend HC3 \cite{guoHowCloseChatGPT2023} by translating its English part to French using Google Translate and add further French out-of-domain (OOD) examples to make models trained on this data more robust. The OOD dataset consists of direct French responses by ChatGPT and BingChat to translated questions from the HC3 dataset (as opposed to translating the answers as done originally), question-answer pairs from the French part of the multi-lingual QA dataset MFAQ \cite{debruynMFAQMultilingualFAQ2021}, and sentences from the French Treebank dataset (Le Monde corpus). Finally, the dataset also contains a small number of adversarial examples written by humans with access to ChatGPT to obtain a similar style to that of ChatGPT.

\subsubsection{\citealt{liaoDifferentiateChatGPTgeneratedHumanwritten2023}}

Focusing on the medical domain, \citet{liaoDifferentiateChatGPTgeneratedHumanwritten2023} build on two public medical datasets: a set of medical abstracts from Kaggle\footnote{\url{https://www.kaggle.com/datasets/chaitanyakck/medical-text}} and radiology reports from the MIMIC-III dataset \cite{johnsonMIMICIIIFreelyAccessible2016}. 
ChatGPT is given parts of an example medical abstract or a radiology report for the machine-generated samples and is prompted to continue writing it.
The authors state that text continuation can generate more human-like text compared to rephrasing or direct prompting. 

\begin{table*}[t!]
    \centering
    \begin{tabular}{lllcc}
        \toprule
        \thead{Paper} & \thead{Dataset} & \thead{Approaches} & \thead{Explainability} & \thead{Code} \\
        \midrule
        \citealt{mitrovicChatGPTHumanDetect2023} & 
            \citealt{mitrovicChatGPTHumanDetect2023} & 
            \makecell[l]{DistilBERT \\PBC} &
            SHAP  & 
            \n \\
            \midrule

        \citealt{liaoDifferentiateChatGPTgeneratedHumanwritten2023} & 
            \citealt{liaoDifferentiateChatGPTgeneratedHumanwritten2023} & 
            \makecell[l]{BERT \\PBC\\XGBoost \\CART} &
             \href{https://github.com/cdpierse/transformers-interpret}{transformer-interpret} & 
            \n \\\midrule

           \citealt{liuArguGPTEvaluatingUnderstanding2023} & 
       \citealt{liuArguGPTEvaluatingUnderstanding2023} (ArguGPT) & 
        \makecell[l]{RoBERTa-large \\SVM} &
        \n  & 
        \href{https://github.com/huhailinguist/ArguGPT}{\y*} \\ \midrule

        \citealt{guoHowCloseChatGPT2023} & 
        \citealt{guoHowCloseChatGPT2023} (HC3) & 
        \makecell[l]{GLTR \\ RoBERTa-single \\ RoBERTa-QA} &
        \n  & 
        \href{https://github.com/Hello-SimpleAI/chatgpt-comparison-detection}{\y} \\ \midrule

         \citealt{antounRobustDetectionLanguage2023} & 
        \makecell[l]{\citealt{antounRobustDetectionLanguage2023}\\ \citealt{guoHowCloseChatGPT2023} (HC3)} & 
        \makecell[l]{CamemBERT  \\ CamemBERTa\\ RoBERTa  \\ ELECTRA \\ XLM-R } &
        \n  & 
        \href{https://gitlab.inria.fr/wantoun/robust-chatgpt-detection}{\y} \\ \midrule

        \citealt{vasilatosHowkGPTInvestigatingDetection2023} & 
        \citealt{ibrahimPerceptionPerformanceDetectability2023} & 
        \makecell[l]{PBC} &
        \n  & 
        \n \\ 
        \bottomrule
    \end{tabular}
    \caption{Methods proposed in the literature for detecting ChatGPT-generated text. PBC: Perplexity-based classifier. Publicly available models can be accessed by clicking on the \y character. *Authors indicate it will be made available at a future date.}
    \label{tab:Methods}
\end{table*}

\subsection{Methods}
 
In this section, we report on the various methods proposed for detecting ChatGPT-generated text. The scope of this review does not include the evaluation or comparison of the results obtained from these methods. This limitation primarily arises from the absence of a common experimental setup and the utilization of different datasets and metrics.
Table \ref{tab:Methods} provides an overview of these recent approaches.

Some previous works have utilized transformer-based models to classify text generated by ChatGPT and human-written text, as demonstrated by \citet{mitrovicChatGPTHumanDetect2023}.
Their approach consists of two components: a detection model and a framework to explain the decisions made by this model.
They first fine-tune an uncased version of DistilBERT \citep{sanhDistilBERTDistilledVersion2019} and then employ SHAP to provide local explanations in the form of feature importance scores to gain insights into the significance of different input features of the model's results.
As a baseline comparison, they implement a perplexity-based classifier that categorizes text based on its perplexity score, where GPT-2 is used for calculating perplexity scores.
Their results show that the DistilBERT-based detector outperforms the perplexity-based classifier.
However, its performance decreases when considering the rephrased dataset by ChatGPT.

In \citet{liaoDifferentiateChatGPTgeneratedHumanwritten2023}, different models are proposed to detect medical text generated by ChatGPT: a fine-tuned BERT model \citep{devlinBERTPretrainingDeep2019}, a model based on Classification and Regression Trees (CART), an XGBoost model \citep{chenXGBoostScalableTree2016} and a perplexity classifier that utilizes BioGPT \citep{luoBioGPTGenerativePretrained2022} for calculating text perplexity. Predictions by the BERT model are explained by visualizing the local features of the samples, where it can be seen that using conjuncts is an essential feature for the model classifying a medical text as machine-generated. 

\citet{liuArguGPTEvaluatingUnderstanding2023} fine-tune RoBERTa to detect argumentative essays generated by different GPT models, including ChatGPT, and evaluate its performance on document, paragraph, and sentence-level classification. The essays are broken down into paragraphs and sentences for paragraph and sentence-level classification. They train and compare the performance of SVM models using different linguistic features. These models serve as a baseline to compare with the RoBERTa model and to understand which linguistic features differentiate between human and ChatGPT-generated text.    

\citet{guoHowCloseChatGPT2023} implement a machine learning and deep learning-based detector. They utilize a logistic regression model trained on the GLTR Test-2 dataset \citep{gehrmannGLTRStatisticalDetection2019} and two deep classifiers based on fine-tuning the pre-trained transformer model RoBERTa. One deep classifier is designed explicitly for single-text detection, while the other is intended for QA detection. The authors construct various training and testing datasets versions to assess the models' robustness.
They create full-text, sentence-level, and mixed subsets of the collected corpus. Each subset has both a raw version and a filtered version where prominent indicating words referring to humans (such as ``Nope'' and ``Hmm'') or ChatGPT words (such as ``AI assistant'') are removed. The evaluation of the models reveals that the RoBERTa-based models outperform GLTR in terms of performance and exhibit more robustness against interference. Moreover, the RoBERTa-based models are not influenced by indicating words.

Building upon the work of \citet{guoHowCloseChatGPT2023}, \citet{antounRobustDetectionLanguage2023} propose an approach for developing robust detectors able to detect  ChatGPT-generated text in different languages, with a focus on French.
Their approach consists of fine-tuning pre-trained transformer-based models on English, French, and multilingual datasets. They train RoBERTa and ELECTRA \citep{clarkELECTRAPretrainingText2020} models on the English dataset,  CamemBERT \citep{martinCamemBERTTastyFrench2020} and CamemBERTa \citep{antounDataEfficientFrenchLanguage2023} on the French datasets and \mbox{XLM-R} \citep{conneauUnsupervisedCrosslingualRepresentation2020} on the combined English and French dataset.
They evaluate the robustness of these models against adversarial attacks, such as replacing characters with homoglyphs and adding misspelled words. Considering in-domain text, their results show that French models perform well in detecting machine-generated text.
Still, they were outperformed by the English models, while \mbox{XLM-R} provides the best and most resilient performance against adversarial attacks for both English and French.
However, this performance decreases when evaluated on out-of-domain text. \\
\indent Another method proposed for detecting ChatGPT-generated text is a metric-based approach proposed by \citet{vasilatosHowkGPTInvestigatingDetection2023} to detect machine-generated student assignments by calculating perplexity scores using GPT-2. They show that having category-wise thresholds (derived from dataset metadata) results in better detection performance than only having one threshold value. 

\subsection{Analysis of Human and ChatGPT-Generated Text}

The textual characteristics of ChatGPT-generated text as well as its syntactic and linguistic features, are of significant focus in the works we reviewed. 
These linguistic and stylistic features are compared to the human-written texts in the datasets.
In this section, we summarize and provide an overview of the findings of such analyses for the different domains and datasets we reviewed.

\begin{itemize}
    \item \textbf{Medical domain:} Medical texts generated by ChatGPT have lower text perplexity and are more fluent, neutral, positive, and logical but more general in content and language style, while medical texts written by humans are more diverse and specific \citep{liaoDifferentiateChatGPTgeneratedHumanwritten2023}.
    \item \textbf{English argumentative essays:} ChatGPT produces syntactically more complex sentences than English language learners, but ChatGPT-authored essays tend to have lower lexical diversity \citep{liuArguGPTEvaluatingUnderstanding2023}.
    \item \textbf{Multi-domain question answering:} ChatGPT writes in an organized and neutral way, offers less bias and harmful information, and refuses to answer questions where it believes it does not know. ChatGPT answers are formal, less emotional, and more objective than human answers \citep{guoHowCloseChatGPT2023}.
    \item \textbf{Scientific abstracts:} ChatGPT has a better choice of vocabulary, can generate more unique words, uses more connecting words, and has fewer grammatical errors \citep{yuCHEATLargescaleDataset2023}.
    \item \textbf{Language-agnostic characteristics:}   The linguistic and syntactic characteristics of ChatGPT-generated text tend to be language-agnostic. Text generated in different languages, such as English, French, and Chinese, shows similar characteristics where ChatGPT tends to produce didactic and impersonal text without errors. Such errors can indicate human text, like grammatical, spelling or punctuation mistakes \citep{antounRobustDetectionLanguage2023, guoHowCloseChatGPT2023}.
\end{itemize}

\subsection{General Insights}

Based on trends and regular mentions we encountered during the creation of our review, we now report some general insights on the state of detecting ChatGPT-generated text.

\paragraph{Role of explainable AI:} Explainability techniques such as SHAP are helpful with detection models. These techniques provide insights into the most important features and words that contribute to classification, thus allowing a better understanding of the writing styles of humans and ChatGPT. This is also valuable in debugging detectors as they can highlight the main words contributing to the misclassification and thus enable better analysis of such models.   

\paragraph{Humans versus ChatGPT in detection task:} Another insight is that humans are worse at detecting machine-generated text by ChatGPT compared to ChatGPT itself. With additional training, humans would achieve better results. 

\paragraph{Robustness of detectors:} The robustness of detectors improves when they are trained on datasets that are extended to include also perturbed data, such as homoglyphs and misspellings. This might help the detectors focus more on writing style than writing errors. When evaluated on out-of-domain texts, the performance of detectors tends to decrease, especially when adversarial text is included.

 \paragraph{Impact of text length on detection:} The shorter the text length, the more challenging and less reliable detection becomes. Models trained on datasets containing full text and question-answer subsets (including answer contexts) do not perform well when evaluated on short texts such as sentences or smaller QA subsets.

\paragraph{Lack of special prompts in ChatGPT-generated text:} Some conclusions and analyses in the reviewed papers have been made based on considering text generated by ChatGPT using its most general style and state, i.e., without using any special prompts that could ask ChatGPT to pretend to be a certain writer or to write in a special style. This could be an interesting area of investigation for future work, where new datasets are constructed, and the robustness of detectors against this type of text is tested. 

\paragraph{Perplexity-based detectors:} perplexity-based detectors depend on using open-source \acp{llm} like GPT-2 and BioGPT to calculate perplexity scores. As ChatGPT generates the target text, calculating these scores using ChatGPT could benefit a lot in this task, as seen with other models using this method. However, this is not possible due to the unfortunate fact of it being a closed-source model.   

\paragraph{Cost of constructing machine-generated datasets:} Constructing and utilizing large-scale ChatGPT-generated datasets is important for drawing more generalized and precise conclusions. Therefore using ChatGPT's API is essential for this sake. However, the costs of doing so can be prohibitive.

\paragraph{Multilinguality:} Our sample of papers has English dominance and performance for other languages is worse. Just as in NLP in general \citep{artetxeCallMoreRigor2020}, we call for more work in this area. This could help explain why some detectors are less reliable in detecting machine-generated text when the text is translated into different languages.

\section{Conclusion and Future Work}

The impressive capabilities of ChatGPT in producing high-quality and convincing text have brought attention to the risks associated with its improper usage across different domains. 
Consequently, the reliable detection of ChatGPT-generated text has become an important task. To address this concern, numerous datasets and detection methods have been proposed. 
In this paper, we provided a concise overview  of the diverse datasets created, proposed methods, and qualitative insights of comparing human-written text with text generated by ChatGPT.

We see a wide variety of approaches and datasets in the papers we discussed.
On the one hand, this is good to see since many factors, such as the domain, language, or format, influence the detection task.
On the other hand, we also see a big diversity in experimental and dataset setups.
Some works use adversarial examples, and others do not.
Some allow the rephrasing of human text by ChatGPT, while others use purely human versus machine-generated text.
Some works include the prompts and ChatGPT versions they used to generate the data; others do not.
These, among other differences, make comparisons difficult, which is one reason we do not include scores in this survey. 
This also highlights important future work, namely to test \emph{methods across datasets} and \emph{datasets across methods}.

Another factor to consider is the domain of the text.
The datasets we have discussed are in diverse domains and cover at least two important ones affected by ChatGPT's risks: health and education.
One notable domain we did not encounter is (fake) news.
Although this is a big \ac{nlp} field on its own, we expected more attention for it in the context of ChatGPT.
Future work can definitely help in this area.
The format of the text is related to the domain and is another important factor to consider.
For example, the shared tasks we mentioned provide tweets, news articles, or reviews as their formats.
A systematic look at format and domain influence concerning ChatGPT could be valuable future work.


Multilinguality is another open problem.
As with virtually all NLP tasks, we have seen that English is, unfortunately, the dominant language in the datasets.
Experiments and gathering datasets across different languages are important future directions.
The current task could also draw inspiration from the field of machine translation.
It has a long and ongoing history of trying to detect (badly) translated text, so-called \emph{translationese} \cite{baroni2006new}, which could be used or adapted to detect general machine-generated text.

Lastly, an important factor we have not seen discussed much is the temporal aspect of ChatGPT.
Outputs might change over time, especially since it is a closed-source system.
This calls for repeated tests over time to ensure detection methods are not regressing in their performance.
Machine-generated text detection is also a cat-and-mouse game; since models are optimized to mimic human language, detection becomes harder and harder.

\section{Limitations}

A limitation of our work is that recent methods proposed for detecting ChatGPT-generated text are pre-prints published in arXiv, due to the rapid pace of work in this area. Additionally, we limit our scope to academic papers and exclude online non-academic tools as we do not know how those tools were trained or how they work internally.

This is also a big problem when discussing ChatGPT in general.
Since it is a closed-sourced system without detailed information about its training and dataset, it is impossible to know if the results are reproducible.
Models can change at any moment in the background, models can be decommissioned, or the price of access can change drastically.
We are well aware of and concerned about these developments, but given the significant opportunities and risks ChatGPT poses, we believe a survey like this one is valuable.

\section*{Acknowledgements}
We thank Florian Matthes and the Software Engineering for Business Information Systems (SEBIS) chair at TUM for their funding and support. We also thank the anonymous reviewers for their helpful and insightful comments.

\bibliographystyle{acl_natbib}
\bibliography{ranlp2023}


\end{document}